# Generation and Interpretation of Temporal Decision Rules

Kamran Karimi and Howard J. Hamilton

Department of Computer Science,
University of Regina,
Regina, Saskatchewan, Canada S4S 0A2
{karimi, hamilton}@cs.uregina.ca

*Abstract:* We present a solution to the problem of understanding a system that produces a sequence of temporally ordered observations. Our solution is based on generating and interpreting a set of temporal decision rules. A temporal decision rule is a decision rule that can be used to predict or retrodict the value of a decision attribute, using condition attributes that are observed at times other than the decision attribute's time of observation. A rule set, consisting of a set of temporal decision rules with the same decision attribute, can be interpreted by our Temporal Investigation Method for Enregistered Record Sequences (TIMERS) to signify an instantaneous, an acausal or a possibly causal relationship between the condition attributes and the decision attribute. We show the effectiveness of our method, by describing a number of experiments with both synthetic and real temporal data.

## 1. Introduction

Broadly speaking, the problem considered in this paper is that of understanding a system that produces data sequentially, in a temporally ordered manner. The data are in the form of a sequence of values for a set of attributes. Measuring and interpreting the influence of the attributes on each other allow us to better understand the system. We assume the influence of certain attributes on others manifests itself within a limited number of consecutive records (a window of records). We consider some attributes to be related if the values of several attributes (*condition attributes*) can be used to predict the value of another one (the *decision attribute*). Predicting the value of a decision attribute is done using a *decision rule*, such as if {(Outlook = sunny) and (Temperature > 20)} then (Play = yes) (Rule 1.1), where Play is a decision attribute, and Outlook and Temperature are condition attributes. A decision rule for predicting the value of a discrete decision attribute is called a *classification rule*. A set of decision rules that predict the same decision attribute is called a *rule set*.

Our approach is based on generating sets of *temporal decision rules*, and using them to interpret the relationships among the attributes. A temporal decision rule (or simply a *temporal rule*) is a decision rule that can be used to predict or retrodict the value of a decision attribute, using condition attributes that are observed at times other than the decision attribute's time of observation. In such a rule the temporal relationships between the condition attributes and the decision attribute are explicitly represented. An example temporal rule is: if {(Outlook$_{t-1}$ = sunny)} then (Outlook$_t$ = sunny) (Rule 1.2), where $t$ represents the current day and $t$ – 1 represents the preceding day. In such a temporal rule, the time of occurrence (represented by the subscript $t$) of the decision attribute is called the *current time*. Extracting such a rule requires an input dataset where observations are made every day and the temporal order can be seen or deduced.

As we will see in this paper, temporal decision rules are useful by themselves, because they may have higher accuracy values than non-temporal ones. However, we take another step: considering that each rule set expresses a relationship between the condition attributes and the decision attribute, we provide a way to interpret that relationship as *instantaneous, acausal*, or *possibly causal*. Causality is a complex subject and its definition is a matter of debate [5, 8, 18, 24, 31, 35]. We consider our definitions in this area to be alternatives to many others.

Non-temporal decision rules, as in Rule 1.1, are well known. We augment this form of rule by considering the temporal nature of the data, where previous or future values of the condition attributes may appear to exert influence on the value of the decision attribute. The past affecting the present follows the normal direction of time. This direction is consistent with our everyday observations of causality. If using the past condition attributes results in better decision making, then a causal relation may exist. Another possibility is that we may be observing a temporal pattern that is not causal. For example, two attribute values may be related to each other over time (first, one is observed, and after a while the other one's value is observed), but neither is causing the other. They may both have a hidden common cause that has escaped our attention. So the observation that a temporally ordered relationship exists between some attributes by itself does not justify the conclusion that causality exists. We refer to any temporal relationship where the past appears to determine the present as a *possibly causal relationship*, which for simplicity of terminology we call a *p-causal relationship*.

A temporal relationship that is not p-causal is said to be *acausal*. To investigate whether a temporal relation is p-causal or acausal, we consider the possibility that the value of the decision attribute is determined by values in preceding records, in succeeding records, or in both preceding and succeeding records. An example rule that depends on both preceding and succeeding records is: if {(Outlook$_{t-1}$ = overcast) and (Outlook$_{t+1}$ = rainy)} then (Outlook$_t$ = rainy). (Rule 1.3). Since we are interested in finding possibly causal rules, identifying some relationships as acausal is valuable, because it suggests that they are unlikely to be causal.

The definitions of causality and acausality given above are widely used in the study of linear systems. For example, in signal processing, the definitions of causal and non-causal filters [22] are similar to the definitions of p-causal and acausal rule sets, respectively.

The unit of time progression depends on the dataset and in particular the frequency at which its records were observed. In Rules 1.2 and 1.3, the unit is a day, reflecting the assumption that the data were gathered daily. As an alternative, one can read "$t+1$" as "next" and "$t-1$" as "previous", and use the same method for any "$t \pm n$" (read: $n$ observations after or before).

In this work, we focus on three mutually exclusive (but not exhaustive) possibilities for a relationship between a set of condition attributes and a decision attribute. First, a relationship is *instantaneous* (or *non-temporal*) if it is best described as a traditional decision rule, and the values of the condition attributes are used to determine the value of the decision attribute at the same instant. Secondly, a relationship is *possibly causal* (or *p-causal*) if only values for condition attributes in preceding records are used to predict the value of the decision attribute in the current record. Thirdly, a relationship is *acausal* if the value for at least one condition attribute from a succeeding record is used to *retrodict*, i.e., predict in reverse, the decision attribute in the current record. In other words, if a relationship involves determining a current decision value based on later values, it is not a causal relationship. Three other possibilities exist, corresponding to combinations of values from the preceding and current records, current and succeeding records, and preceding, current, and succeeding records, but we have not found it useful to distinguish them. These concepts are introduced formally in Section 3.

The Temporal Investigation Method for Enregistered Record Sequences (TIMERS), designed by the authors and implemented in a software package called TimeSleuth [12], generates rule sets and measures their accuracy values. TIMERS begins by pre-processing the temporal input data to make them compatible with conventional (non-temporal) decision rule generators. After that, rules are generated and their quality, as measured by their training or predictive accuracy, is used to give a verdict on the causal or acausal nature of the relationship expressed by the rules. By default, TimeSleuth uses a modified version of C4.5 [25] for rule generation. Our modifications to C4.5 allow it to function seamlessly with TimeSleuth, as explained in [10]. These modifications also enable C4.5 to convert decision rules into PROLOG statements [9]. As shown in [13], the TIMERS algorithm can also use regression software, such as CART [3], for rule generation.

The algorithm depends on a rule generator, and the expressive power of the output, as well as the training and predictive accuracy of the rules depend on this generator. C4.5, for example, uses if-then rules, and any causal relation must be representable using this format if it is to be discovered by TIMERS.

This work presents two main contributions. (1) We introduce a method to extract temporal decision rules using existing rule discovery software. The discovered decision rules can predict the value of a decision attribute before or after the values of condition attributes are observed, making these rules useful for predicting future behaviour and also applications such as recovering missing or faulty values in a dataset. (2) We propose a link between the ability to predict or retrodict values on the one hand and acausality or possible causality on the other.

The remainder of this paper is organised as follows. Section 2 describes previous work done in temporal and causal information discovery. Section 3 formally presents the problem and provides algorithms for solving it. Section 4 presents the results of experiments with the TIMERS method. Section 5 concludes the paper and discusses future work.

## 2. Background in Temporal and Causal Rule Discovery

In this section, we describe previous work on extracting temporal or causal information from data. What sets our work apart from the previous research is the choice of problem domain. We choose to say that "a relationship is describable among the attributes" if and only if at least one set of decision rules involving these attributes can be created that has predictive accuracy above a specified threshold. This approach is well suited for temporal data with discrete decision attributes. It is poorly suited to many univariate and multivariate time series with continuous valued decision attributes. Also, we augment the concept of a decision rules by allowing retrodiction of the value of the decision attribute, which is a novel extension that can be useful when applied to accumulated datasets in an offline manner.

### 2.1. Temporal Discovery

Temporal data are often represented as a sequence, sorted in a temporal order. Examples of studies of sequential data and sequential rules are given in [1, 7, 28]. There are a number of general fields in the study of sequential data. A *time series* is a time-ordered sequence of observations taken over time [2, 4]. An example is the series of numbers <1, 3.5, 2, 1.7, …>. In a *univariate time series*, each observation consists of a value for a single attribute, while in a *multivariate time series*, each observation consists of values for several attributes. Most research on time series has assumed the presence of a distinguished attribute representing time, and numeric values for all other attributes. Attempts have been made to fit constant or time-varying mathematical functions to time series data [2]. A time series can be regular or irregular, where a *regular time series* consists of data collected at predefined intervals. An *irregular time series* consists of data collected in order, but at varying intervals of time. A *deterministic time series* can be predicted exactly, while the future values in a *stochastic time series* can only be determined probabilistically [4]. The former is a characteristic of many artificial systems, while the latter applies to most natural systems. Simple operations like determining the minimum or maximum values of certain attributes, finding trends (such as increases or decreases in the value of stocks), cyclic patterns (such as seasonal changes in the price of commodities), and forecasting are common applications of time series data.

In [23], the phrase *multiple streams of data* is used to describe simultaneous observations of a set of attributes. The streams of data may come from different sensors of a robot, or the monitors in an intensive care unit, for example. The values in these streams are recorded at the same time, and thus form a multivariate time series. The data represented in Table 1 is an example of multiple streams of data, where three attributes are observed over time. In [21], an algorithm is presented that can find rules (called "structures" by its authors) relating the previous observations to the future observations. Such temporal data appear in many application areas; a good overview of these applications can be found in [26].

| Outlook | Temperature | Play |
|---------|-------------|------|
| Sunny | 25 | Yes |
| Rainy | 13 | No |
| Overcast | 20 | Yes |
| Sunny | 10 | No |

**Table 1.** Records containing values of 3 attributes

An *event sequence* is a series of temporally ordered events, with either an ordinal time attribute (which gives the order but not a real-valued time) or no time attribute. Each event specifies the values for a set of attributes. A recurring pattern in an event sequence is called a *frequent episode* [20]. Recent research has emphasised finding frequent episodes with varying number of events between the key events that identify the event sequence. Algorithms such as Dynamic Time Warping and its variants measure the similarity of patterns that are stretched differently over time [17]. These methods have apparently not been applied to searching for causal relations in data, and no claims have been made as to whether or not they represent causal relationships. Henceforth, we restrict the term "time series" to refer to a sequence of sets of continuous (real) values, while an event sequence can contain values drawn from symbolic domains.

In [36], Zhang describes an algorithm for detecting temporal relationships between time-stamped events. Instead of using fixed window sizes, this method uses maximum and minimum time lag values to create rules that associate two events. The rules are then evaluated using a number of metrics (including accuracy) and the best ones are presented as the output.

Despite having different terminology, all the representations described so far in this section have the common characteristic of recording the values of some attributes and placing them together in a record. Time series, event sequences, and streams of data all can be used as input when searching for temporal rules (called *patterns*, *episodes*, and *structures*, respectively).

*2.2. Causal Discovery*

Here we review a few key terms related to causality. We also describe several software applications that implement previous approaches to automatically detecting causal relationships.

Causality usually implies cause(s) appearing before the effect(s). An effect happening before its cause is called *backward causation*. As shown in [32], some philosophers do not consider this impossible or paradoxical. In this paper, however, we assume that backward causation is not possible. We refer to apparent backward causation as acausality, and we assume that it implies the presence of a hidden common cause. The effects of such a hidden common cause are spread over time, and if only the effects are observed, backward causation may appear to exist. This interpretation of the philosophical notion of backward causation acknowledges the temporal characteristics of the relationship, and avoids any temporal paradox.

One definition of causality is named *Granger causality*, which is used mainly in economic contexts; this definition puts clear emphasis on temporal precedence [6]. A *Granger causal relationship* exists when using previous values of some attributes improve the prediction of a decision attribute's value. Suppose we are observing two attributes $x_t$ and $y_t$ over time, and $A$ is a set of attributes considered relevant to $y_t$. We say $x_t$ *granger-causes* $y_t$ if there is a natural number $h > 0$ such that $P(y_{t+h} | A) < P(y_{t+h} | A \cup \{x_t, x_{t-1}, \ldots, x_1\})$, where $P(a | b)$ is the probability of event $a$ happening, given that event $b$ has happened [35].

*Probabilistic causality* can be defined as follows. $A$ is a *probable cause* of $B$ if $P(B | A) = 1$ and $P(B | \sim A) = 0$. This definition implies temporal precedence of the probable cause with regards to the effect. In practice, however, this definition is too brittle, and few real-world datasets satisfy it because of noise in data and complicated causal relationships. Instead, one says $A$ is a *probable cause* of $B$ if $P(B | A) > P(B | \sim A)$. Both definitions implicitly assume that no common cause is present. For example, let $A$ = seeing lightning and $B$ = hearing thunder. If we hear thunder only after seeing lightning, then we could conclude $P(B | A) > P(B | \sim A)$, which would lead us to believe that the act of seeing the lightning causes the hearing of thunder. For this reason, in the probabilistic approach, an assumption is made that no hidden common causes are at work. In this example, the lighting and the thunder are created at the same time by a discharge of electricity, which is the hidden cause of both phenomena.

Not every conditional probability implies a temporal ordering. In our description of the statement $P(B | A)$, event $A$ was assumed to have already happened, and this temporal ordering should not be ignored. Bayes' rule states $P(B | A) = P(A | B) \times P(B) / P(A)$. However, this rule should not be applied to conditional probabilities representing causal relations without considering the temporal order of events. In the left hand side of the equation, event $B$ is assumed to have already happened, while in the right hand side event $A$ is assumed to have happened. In other words, careless algebraic manipulation could disturb information about the order in which the events are believed to have happened.

Automatic discovery of causal relations among a number of attributes has been an active research field. More specifically, automated methods have been applied to determining whether or not the value of an attribute is caused by the values of the condition attributes. The main approach to the computational discovery of causality has used the notion of conditional independence and the creation of Bayesian networks [24]. TETRAD [29] and CaMML [16, 33] are examples of applications that work on this basis. As shown in [14], in domains where the value of an attribute can be reliably determined by other attributes, TIMERS performs better than CaMML and TETRAD.

The Bayesian approach represented by Pearl's work is described in [24]. A Bayesian network uses the statistical concept of conditional independence as a measure of the control that one attribute may have over another [24]. For example, given the three attributes $x$, $y$, and $z$, if $y$ is independent from $z$ given $x$, that is, $P(y, z | x) = P(y | x)$, then Pearl concludes that $y$ is not a direct cause of $z$. In other words, $x$ *separates* $y$ and $z$ from each other. In the corresponding Bayesian network, no edge exists between $y$ and $z$. Bayesian networks show the conditional dependence of attributes as edges. For example, $P(y, z | x) = P(y | x)$ is represented as $y$ — $x$ — $z$. Discovering the direction of each edge needs further investigation. Other representations that are similar to Bayesian networks are belief networks, probabilistic networks, knowledge maps, and causal networks [27]. In a *causal Bayesian network*, the edges between variables are interpreted as signifying causal relations [34].

## 3. Discovering Temporal Rules and Causality

In this section, we explain how TIMERS works. First our definitions and assumptions are presented, and then the algorithms are introduced for discovering temporal rules and making a judgement about whether a relationship is instantaneous, acausal or p-causal.

### 3.1. Definitions and Assumptions

In the following formal definitions of instantaneous, acausal, and p-causal rule sets, let **D** be a set of input records defined over the set of input attributes $A = \{a_1, ... , a_m\}$. It is assumed that the input is an event sequence. Let $w$ be the window size, which is the number of consecutive records that are merged together in an operation called temporalisation, as explained in Section 3.2. Let $R = \{r_1, ..., r_n\}$ be a rule set generated from **D**, using a window size of $w$, to predict the value of a decision attribute. DECISION() is an operator that, given a decision rule, returns the corresponding decision attribute. All rules in $R$ have the same decision attribute, called $d_{t0}$, and thus, $\forall i, 1 \leq i \leq n$, DECISION$(r_i) = d_{t0}$, where $d \in A$ is the name of an input attribute, and $t_0$ is the time of its occurrence with respect to the window, with $1 \leq t_0 \leq w$. CONDITIONS$(r)$ is the set of all condition attributes used in rule $r$. Attributes in CONDITIONS$(r)$ have a time step as index, to distinguish the values for the same attribute observed at different times. For example, $a_t \in$ CONDITIONS$(r)$ means that attribute $a$ has been observed at time step $t$, which may or may not be the same as time step $t_0$ (when the decision attribute was observed). It does not make sense to use the attribute $d_{t0}$ to predict itself at the same time. For any attribute $a_t$ in CONDITIONS$(r)$, we have either $a \in A - \{d\}$ and $1 \leq t \leq w$ (condition attributes can appear at any time step) or $a = d$ and $t \neq t_0$ (the decision attribute can appear at any time step other than $t_0$).

We assume that DECISION$(r) \notin$ CONDITIONS$(r)$. In other words, an attribute at a given time cannot be used to predict its own value at that time. Here we are concerned with the properties of sets of rules, rather than attributes, and we assume the rule sets not to be empty, that is, we assume that for any rule set $R$, there is at least one $r \in R$ such that CONDITIONS$(r) \neq \varnothing$.

**Instantaneous**. An instantaneous rule set is a set of rules such that in every rule the current value of the decision attribute is determined only by the values of the condition attributes observed at the same time step as the decision attribute.

**Definition 1:** A rule set $R$ is *instantaneous* iff for all $r \in R$, if DECISION$(r) = d_{t0}$, then for all $a_t \in$ CONDITIONS$(r)$, $t = t_0$.

**Temporal**. A temporal rule set is a set of rules where the condition attributes appear at different time steps than the decision attribute. We exclude any condition attribute that appears at the same time as the decision attribute so as to prevent a primarily instantaneous relationship from appearing as a temporal relationship.

**Definition 2:** A rule set $R$ is *temporal* iff for all $r \in R$, DECISION$(r) = d_{t0}$, then for all $a_t \in$ CONDITIONS$(r)$, $t \neq t_0$.

A temporal set of rules can be either p-causal or acausal.

**Causal.** In a p-causal rule set, the current value of the decision attribute relies only on the previous values of the condition attributes in each rule [30].

**Definition 3:** A rule set $R$ is *p-causal* iff for all $r \in R$, DECISION$(r) = d_{t0}$, then for all $a_t \in$ CONDITIONS$(r)$, $t < t_0$.

**Acausal.** In an acausal rule set, the current value of the decision attribute relies on the future value of at least one condition attribute [19].

**Definition 4:** A rule set $R$ is *acausal* iff (1) for all $r \in R$, if DECISION($r$) = $d_{t_0}$, then for all $a_t \in$ CONDITION($r$), $t \neq t_0$, and (2) if there exists $r \in R$ where DECISION($r$) = $d_{t_0}$, then there also exists $a_t \in$ CONDITIONS($r$), $t > t0$.

Two assumptions are made. First, we assume that the predictive accuracy of the rules is a suitable measure of their quality. The training accuracy of the rules can be substituted if a user prefers to do so. Second, we assume that there is an order of conceptual simplicity among the three types of the relations, with instantaneous being the simplest type of relationship, followed by acausality, and then p-causality. Hence, instantaneous $<_{simplicity}$ acausal $<_{simplicity}$ p-causal. The intuition behind this ordering is that as we move from instantaneous to acausal and then to p-causal, more claims are being made about the relationship. As a principle, we try to explain a relationship with the simplest possible type. This ordering is used to choose a winning relation type when the results of the tests for each of the three relation types are close. As a simple example, suppose one attribute records the water level in a lake every hour and two other attributes tell whether it is raining 15 km and 30 km upriver from the lake. If all observations were from a time when the water level was constant, the relationship would be identified as instantaneous, because this explanation is simplest according to our stated assumption of the order of conceptual simplicity, given in Section 3.1. If the water level was steadily decreasing and no rain was reported, a temporal relationship between the current water level and either the preceding or the succeeding water level could be identified. It would be described as acausal because it can be predicted equally effectively using succeeding or preceding water levels. If the water level only increased one hour after rainstorms 15 km upriver from the lake or two hours after rainstorms 30 km upriver, then the relationship would be identified as p-causal.

*3.2. Temporalisation*

To combine data from multiple time steps, a variety of windowing techniques are used in time series analysis and other types of temporal analysis [39]. Due to its importance in the TIMERS method, we specify our technique, which we call *temporalisation*, in some detail.

The temporalisation technique prepares the data for rule extraction, and the final judgment about the type of the relationship is based on the quality of the rules. The quality can be measured using the accuracy of the rules. For the instantaneous test, no temporalisation is performed. Alternatively, one could say we temporalise with a window size of 1. For the p-causal (forward in time) test with a window size $w$, the temporalisation involves merging every $w$ consecutive records together, and setting the decision attribute to be in the last record (past values predicting the current time). For the acausality test, $w$ consecutive records are identified, and a series of $w-1$ separate tests are performed, using every position in the window besides the last one as the position of the decision attribute (using the last position constitutes a p-causal test). The term "temporalisation" is used because the temporal information is implicitly stored in the position of the attributes in the resulting merged record. As a result, using a window size $w$, the input is transformed from $n$ records, each containing $m$ attribute values, to $n - w + 1$ records, each containing $(w - 1)m + 1$ attribute values.

With any fixed window size $w$, the *sliding position temporalisation algorithm* performs both p-causal and acausal tests. It first extracts the value of the decision attribute from position 1 of the window, and uses the next $w-1$ records to predict its value. Then the decision attribute is extracted from position 2 of the window, and the previous record (position one) and the next $w$-2 records are used for prediction. This movement of the position of the decision attribute continues until at the end it is set to position *w*-1, and the previous *w*-2 records and the next record are used for prediction.

As an example, consider four temporally consecutive records, each with four fields: $R_1$: <1, 2, 4, true>, $R_2$: <2, 3, 5, true>, $R_3$: <6, 7, 8, false>, and $R_4$: <5, 2, 3, true>. To predict the value of the last attribute using a window size of 3, the records are first merged as shown in Table 2. The value of the decision attribute is indicated in boldface. For the record that includes the value for the decision attribute, none of the condition attributes are considered [11]. The *record.value* notation explicitly shows the value for the decision attribute and the record from which it was extracted. For example, <$R_1$, $R_2$, $R_3$.**false**> corresponds to <1, 2, 4, true, 2, 3, 5, true, **false**>, where **false** is the value of the decision attribute in $R_3$. We put the decision attribute at the end of the temporalised record as a matter of convention.

The second column of Table 2 (sliding position temporalisation) shows the results of using sliding position temporalisation with the decision value extracted from the first position in the window. Similarly, the third column shows the corresponding records when the decision value is extracted from the second position in the window. The fourth column of Table 2 (forward temporalisation) uses only preceding records for prediction. It is equivalent to the sliding position temporalisation with position = 3.

| Instantaneous. *window* = 1 (original data) | Sliding position. *window* = 3, position = 1 | Sliding position. *window* = 3, position = 2 | Forward (Causality). *w* = 3, position = 3 |
|---|---|---|---|
| $R_1$ = <1, 2, 4, **true**> | <$R_2$, $R_3$, $R_1$.**true**> | <$R_1$, $R_3$, $R_2$.**true**> | <$R_1$, $R_2$, $R_3$.**false**> |
| $R_2$ = <2, 3, 5, **true**> | <$R_3$, $R_4$, $R_2$.**true**> | <$R_2$, $R_4$, $R_3$.**false**> | <$R_2$, $R_3$, $R_4$.**true**> |
| $R_3$ = <6, 7, 8, **false**> | | | |
| $R_4$ = <5, 2, 3, **true**> | | | |

**Table 2.** Temporalisation with instantaneous, forward, and sliding position methods

Column 2 of Table 2 shows the records to be apparently disordered, for example the notation <$R_2$, $R_3$, $R_1$.true> is used instead of <$R_3$, $R_2$, $R_1$.true>. The order shown matches the order produced by the sliding position temporalisation algorithm described later in this section. It follows naturally from moving the time of the decision attribute from 1 to 3 and putting the other records in the order of occurrence. With a window size of 3, the records are $R_1$, $R_2$, and $R_3$. At time step 1, the decision attribute is at $R_1$, so the record is shown as <$R_2$, $R_3$, $R_1$.true>, with $R_1$.true at the end to follow the usual convention of placing the decision value last. At time step 2, the record is <$R_1$, $R_3$, $R_2$.true>, etc. At the implementation level, the order of the records included in a temporalised record is not important (e.g., to C4.5). In the temporalised record, each attribute value has a time-step index that determines its time of observation, regardless of where in the merged record it appears. Every output rule is later reordered so that the attributes appear according to their time index, with no temporal ordering violated.

After temporalisation, temporal relations among the merged records are not represented explicitly. There is, however, an implicit temporal relationship within the fields in each temporalised record. A traditional tree or rule generator considers all the attributes to be available at the same time, and temporalisation allows such rule generators to be applied to temporal data, though the output of a traditional tree or rule generator does not consider the attributes to have been observed at different times. After rule generation, the attributes in the output rules can be rearranged according to their time of appearance. We could thus re-temporalise the output rules.

For the acausal test, we can have values from a mixture of preceding and succeeding records. Given a window size *w*, one record contains the value of the decision attribute, *p* records precede this record, and *f* records follow it, i.e., $p + 1 + f = w$. Since the acausality test requires $f$ be at least 1 (at least one value of an attribute from a following record), two inequalities apply: $1 \leq f \leq w - 1$ and $0 \leq p \leq w - 2$. The decision attribute moves from being in the first position (no preceding records) to being in record number $w - 1$ (corresponding to $w - 2$ preceding records and 1 succeeding record). In a p-causality test the decision attribute only appears at position $w$.

```
Temporalise (w, pos, D, d) {
  for (i = 0 to |D| - w)
      temporalisedRecord = < >   //initialise to empty
      for (j = 1 to pos - 1) // previous records
          temporalisedRecord += D_{i+j}
      for (j = pos + 1 to w) // next records
          temporalisedRecord += D_{i+}
      temporalisedRecord += Field(d, D_{i+pos})   // the decision attribute
      output(temporalisedRecord
}
```

**Figure 1.** Sliding position temporalisation method

The algorithm for the sliding position temporalisation operator Temporalise() is presented in Figure 1. The Temporalise() operator takes as input a window size $w$, the position of the decision attribute within the window $pos$, the set of input records $D$, and the decision attribute $d$, and outputs temporalised records. $D_i$ returns the $i$th record in the input $D$. *Field*() returns a single field in a record, as specified by its first parameter. The += operator stands for concatenating the left hand side with the right hand side, with the results going to the left hand side, and < > denotes an empty record.

*3.3. The TIMERS Algorithm*

To use TIMERS, the user chooses a number of input attributes from a dataset $D$. They include the decision attribute in which the user is interested, $d$, and also the condition attributes that will be used for classification. The other attributes present in the data will then be ignored. The input data does not need to be temporalised for testing the instantaneous case, but for both causal and acausal cases the algorithm needs a window size. Since a user may have difficulty determining the most appropriate window size value, TIMERS accepts a range of values, from a minimum window size α to a maximum window size β.

The condition attributes may not have a significant relationship with the decision attribute. In such cases, the decision rules will probably be of low quality. To detect such cases, the user provides a threshold accuracy value $ac_{th}$. The accuracy values of all rule sets produced with various settings are compared to this threshold value, and if all accuracy values are lower, then the algorithm refrains from making any judgment.

TIMERS performs the appropriate temporalisation, generates classification rules, and saves the best accuracy values for each of the p-causal and acausal tests. To provide information for selecting the best method for describing the relationship between the decision and condition attributes, the size of each rule set is saved along with the corresponding accuracy value. The size of a rule set is assumed to partially describe its simplicity.

After this phase, TIMERS decides on the best relationship type. The TIMERS algorithm appears in Figure 2 below.

**Input:** A sequence of sequentially ordered data records $D$, minimum and maximum temporalisation window sizes $\alpha$ and $\beta$, where $0 < \alpha \leq \beta$, a minimum accuracy threshold $ac_{th}$, a decision attribute $d$, and a confidence level $cl$. The attribute $d$ can be set to any of the observable attributes in the system, or the algorithm can be tried on all attributes in turn. *Preference* determines whether the user prefers higher accuracy or a simpler method.

**Output:** A set of accuracy values and a verdict as to the nature of the relationship between the decision attribute and the condition attributes. It could be instantaneous, acausal, or p-causal.

**RuleGenerator()** is a function that receives input records, generates decision trees, rules, or any other representation for predicting the decision attribute, and returns the training or predictive accuracy, as well as the number of rules generated.

**TIMERS(*D*, $\alpha$, $\beta$, $ac_{th}$, *d*, *cl*, preference)** {
 // the instantaneous rule set; window size = 1
 *InstantaneousRS* = RuleGenerator(*D*, *d*)
 *TemporalRS* = Ø // acausal and p-causal rule sets
 **for** (*w* = $\alpha$ to $\beta$) //window range
    **for** (*pos* = 1 to *w*) // decision attribute position
       *TemporalRS* += RuleGenerator(*Temporalise*(*w*, *pos*, *D*, *d*), *d*)

 // at least one condition attribute after the decision
 *AcausalRS* = *SelectAcausal*(*TemporalRS*)

 // condition attributes before the decision attribute
 *PcausalRS* = *SelectPcausal*(*TemporalRS*)

 // get the accuracy of the single rule set
 ($ac_i$, $ruleSize_i$) = max_accuracy(*InstantaneousRS*)

 // best acausal result
 ($ac_a$, $ruleSize_a$) = max_accuracy(*AcausalRS*)

 // best p-causal result
 ($ac_p$, $ruleSize_p$) = max_accuracy(*PcausalRS*)

 // If there is enough relevant information, then stop
 **if** (max($ac_i$, $ac_p$, $ac_a$) < $ac_{th}$) **then** return "No verdict"

 Verdict = "for attribute " + *d* + ", "
 Relation = RelationType(*cl*, ($ac_i$, $ruleSize_i$), ($ac_a$, $ruleSize_a$),
                                    ($ac_p$, $ruleSize_p$), *preference*)
 **case** relation **of**
     INSTANTANEOUS: verdict += "the relation is instantaneous"
     ACAUSAL: verdict += "the relation is acausal"
     P-CAUSAL: verdict += "the relation is p-causal"

 **return** verdict.
}

**Figure 2.** The TIMERS algorithm

The memory space needed by TIMERS is computed as follows. For every run of the Temporalise() operator, we get a dataset of $|D| - (w - 1)$ records. Hence, the total number of output records created by the TIMERS algorithm is $\sum_{w=\alpha}^{\beta} |D| - (w - 1)$. For a window size of 1, the dataset already exists (the original dataset). Temporalised datasets are not kept after being used for rule generation. Thus a maximum of $|D| - (\beta - 1)$ temporalised records are present as input during any iteration. Considering that the number of attributes in each record is multiplied by the window size, the maximum number of fields in the temporalised dataset is $(\beta - 1) \times (|D| - (\beta - 1)) \times m + 1$, where $m$ is the number of fields in each original input data record. The expression $(\beta - 1)$ reflects the fact that for the current time we do not include any of the condition attributes. The $+ 1$ at the end of the formula reflects the decision attribute at the current time.

The number of times that RuleGenerator() runs is equal to $1 + \sum_{w=\alpha}^{\beta} w = 1 + [\beta \times (\beta + 1) - (\alpha - 1) \times \alpha] / 2$. Hence, the time complexity of TIMERS is polynomially related to the time complexity of the RuleGenerator().

TIMERS uses a statistical test to see if the results of the three tests are close together. If so, the simpler method is suggested. As we will explain later, simplicity depends on both conceptual simplicity (for example, an instantaneous relationship is conceptually simpler than a p-causal relationship) and also the number of rules needed to express a relationship (lower number of rules denotes a simpler relationship).

When the accuracy values are close together, users can employ their discretion in choosing a relation type. However, TIMERS uses a statistical method to recommend a relationship type. Using the confidence level provided by the user in the *cl* parameter, it constructs a confidence interval for the accuracy. After computing the interval, it checks to see if the corresponding intervals overlap. If they do, the method with the simpler type of relationship is chosen. The intuition is that even if the simpler method has resulted in less accuracy, it could have *potentially* produced better or the same results.

As an example, suppose with a confidence level of 90%, we have the following accuracy and intervals: the instantaneous accuracy $ac_i = 32.5\%$, $interval_{aci} = [31\%, 34\%]$, the acausal accuracy $ac_a = 35\%$, $interval_{aca} = [33\%, 37\%]$, and the p-causal accuracy $ac_p = 37\%$, $interval_{acp} = [35\%, 39\%]$. Suppose also that all methods resulted in the same number of rules. Because the confidence intervals of the instantaneous method and the acausal methods intersect, instantaneous is chosen because it is considered a simpler explanation that can potentially out perform the acausal case. Then we consider the p-causal case, and since the intervals of the instantaneous and p-causal methods do not overlap, the p-causal method is chosen as the final verdict, because of its higher accuracy value. Figure 3 shows how the best method is selected.

```
Input: A confidence level cl, three accuracy values corresponding to the instantaneous, acausal,
and p-causal methods: $ac_i$, $ac_a$, $ac_p$, and their corresponding number of rules: $nRule_i$, $nRule_a$,
$nRule_p$, a bias b for higher accuracy vs. a simpler method.
Output: A verdict as to the best relationship type.

//info[].method contains one of INSTANTANEOUS, ACAUSAL,
// or P-CAUSAL.
//info[].Accuracy is the best accuracy value.
//info[].interval contains the interval of the accuracy value, computed
// sing a confidence value

Function RelationType(cl, ($ac_i$, $ruleSize_i$), ($ac_a$, $ruleSize_a$), ($ac_p$, $ruleSize_p$), b)
{
  // initialise the info[] structure
  // i = INSTANTANEOUS, a = ACAUSAL, p = P-CAUSAL
  foreach (method = i, a, p)
     info[method] = (method, $accuracy_{method}$, $ruleSize_{method}$, $Interval_{method}$ =
                                  ComputeAccuracyInterval($accuracy_{method}$))
  // if preference is given to higher accuracy, then
  // start the search from lower accuracy values
  if (b = HIGHER_ACCURACY) then
     sort_Ascending(info[])  // sort in ascending order of accuracy.
  else   // SIMPLER_METHOD
     sort_Descending(info[])

  winner = 1
  for (count = 2 to 3)
     if (overlap(info[winner].interval, info[count].interval))  {
       // if there is an overlap, then choose the simpler method
        if (info[count].method $<_{simplicity}$ info[winner].method and
            info[count].ruleSize ≤ info[winner].ruleSize) then
          winner = count
     }
     else { // if no overlap, choose the method with higher accuracy
        if (info[count].accuracy > info[winner].accuracy) then
           winner = count
     }
  //return one of  INSTANTANEOUS, ACAUSAL,
  // or P-CAUSAL
  return info[winner].method
}
```

**Figure 3.** Selecting the best relationship

If needed, the algorithm in Figure 3 could be changed to select the window size that gives the highest accuracy values obtained in either the acausal or p-casual case. In such a case, the order of simplicity could be determined by the window size, with smaller window sizes being considered simpler.

## 4. Experimental Results

In this section, we apply TIMERS to the problem of rule discovery from two temporal datasets. We compare the effects of temporal rule discovery to that of standard C4.5's results, which are obtained when we set the temporalisation window to 1.

The first dataset is from a discrete event simulator called URAL [38], where known atemporal and temporal rules govern an artificial environment. The second dataset is from a weather station in Louisiana AgriClimatic Information System [37].

### 4.1. The Artificial Robot

The world in URAL is a rectangular, 8 × 8 board with a robot moving around in it. *x* (horizontal) values increase towards the left of the screen, and *y* (vertical) values increase towards the bottom. The robot performs a random walk in the domain: at each time-step, it randomly chooses one of the following actions *a*: left (L), right (R), up (U), or down (D). Left and right correspond to moving along the *x*-axis and up and down to moving along the *y*-axis. We used 2500 records for training, and 500 for testing the rules (predictive accuracy). The decision attribute is set to be the current value of *x*, and the other attributes (*y*, and *a*) are set as the condition attributes. There is no relationship between the current value of *x* on one hand, and the current values of *y*, or the direction of the movement, on the other hand. So we predict that an instantaneous test (window size of 1) will give poor results. From our understanding of the domain we know that the current value of *x* depends on the previous value of *x*, and the previous direction of movement. This forward moving relation is deterministic and holds in 100% of the cases. The same holds for *y*. We expect the method to classify the relationship as a p-causal one and assign it a high accuracy score. The acausal hypothesis for *x* says that one can predict where the robot was if one knows where the robot is now. This hypothesis is clearly wrong, as the robot could have arrived at the current position from several previous positions. Hence, we do not expect to get good results with our acausality test when dealing with backward temporal data. The results agree with our expectations and are shown in Table 3, where "T Acc" stands for training accuracy, and "P Acc" stands for predictive accuracy. The "Pos" column shows the index of the decision attribute within the temporalised record.

| Win | Pos | T Acc | P Acc | Type of test | Actual rules |
|---|---|---|---|---|---|
| 1 | 1 | 19.7% | 20.4% | Instantaneous | Instantaneous |
| 2 | 1 | 56.2 | 55.7% | acausal | Acausal |
| 2 | 2 | 100% | 100% | p-causal | p-causal |
| 3 | 1 | 57.6% | 55.6% | acausal | acausal |
| 3 | 2 | 100% | 100% | acausal | p-causal |
| 3 | 3 | 100% | 100% | p-causal | p-causal |
| 4 | 1 | 58.4% | 58.1% | acausal | acausal |
| 4 | 2 | 100% | 100% | acausal | p-causal |
| 4 | 3 | 100% | 100% | acausal | p-causal |
| 4 | 4 | 100% | 100% | p-causal | p-causal |
| 5 | 1 | 58.4% | 57.1% | acausal | acausal |
| 5 | 2 | 100% | 100% | acausal | p-causal |
| 5 | 3 | 100% | 100% | acausal | p-causal |
| 5 | 4 | 100% | 100% | acausal | p-causal |
| 5 | 5 | 100% | 100% | p-causal | p-causal |

**Table 3.** TIMERS' results with robot data: Verdict is causal

There is a clear improvement in accuracy values between temporal rules ($w > 1$) and instantaneous ones ($w = 1$), corresponding to plain C4.5. Considering the 100% accuracy values with a window size of 2 or larger in the p-causal tests,

TIMERS declares the relation to be p-causal. However, the acausal test also gives 100% accuracy for several data sets. With any position greater than 1, the previous record (that contains sufficient information to make an accurate prediction of the current $x$ value), is included in the temporalised data. So TIMERS discovers the correct temporal relation between the current value of $x$ and the previous value of $x$ and movement direction. An example rule is: if $\{(x_1 = 1)$ AND $(a_1 = $ Right$)\}$ then $(x_2 = 2)$. In other words, even with an acausality test, the rules are all p-causal because they only contain attributes from the previous time step. TIMERS checks for this condition and declares a relationship as p-causal even if the corresponding test has been for discovering an acausal relationship.

*4.2. The Louisiana weather data*

The Louisiana weather dataset is a real-world dataset from weather observations in Louisiana. It contains observations of 8 environmental attributes gathered hourly from 22/7/2001 to 6/8/2001. All attributes are recorded as integers. There are 343 training records, each with the air temperature, the soil temperature, humidity, wind speed and direction and solar radiation, gathered hourly. 38 other records were used for testing the rules and generating predictive accuracy values. We have set the soil temperature to be the decision attribute. Since this dataset describes real phenomena, interpreting the dependencies and relationships in it is harder than for the robot dataset. The results obtained are shown in Table 4.

| Win | Pos | T Acc | P Acc | Type of test | Actual rules |
|---|---|---|---|---|---|
| 1 | 1 | 27.7% | 23.7% | Instantaneous | Instantaneous |
| 2 | 1 | 75.1% | 59.5% | Acausal | Acausal |
| 2 | 2 | 82.7% | 67.6% | Causal | Causal |
| 3 | 1 | 85.3% | 75.0% | Acausal | Acausal |
| 3 | 2 | 82.4% | 72.7% | Acausal | Acausal |
| 3 | 3 | 86.8% | 77.8% | Causal | Causal |
| 4 | 1 | 85.3% | 74.3% | Acausal | Acausal |
| 4 | 2 | 85.9% | 74.3% | Acausal | Acausal |
| 4 | 3 | 83.2% | 74.3% | Acausal | Acausal |
| 4 | 4 | 84.4% | 71.4% | Causal | Causal |
| 5 | 1 | 85.0% | 73.5% | Acausal | Acausal |
| 5 | 2 | 87.0% | 76.5% | Acausal | Acausal |
| 5 | 3 | 85.0% | 76.5% | Acausal | Acausal |
| 5 | 4 | 83.8% | 76.5% | Acausal | Acausal |
| 5 | 5 | 86.7% | 73.5% | Causal | Causal |

**Table 4.** TIMERS' results with the weather data: Verdict is acausal

As with the robot dataset, accuracy values are significantly higher with temporalisation ($w > 1$) than with a window size of 1 ($w = 1$). Thus, the relationship is not instantaneous. Because the accuracy values in the causal and acausal tests are close, TIMERS declares the relationship between the other attributes and the soil temperature attribute to be acausal. The relationship is considered acausal because it is temporal but the direction does not matter (the rule set that predicts the current temperature from the previous values of the attributes is approximately as accurate as the rule set that retrodicts the current temperature from the future values).

The verdict that TIMERS gives for a dataset depends on the dataset itself and the performance of the underlying rule generator. If more data were added to the dataset, then the rule sets and thus the verdict could be different. Similarly, if C4.5 were replaced by another rule generator, the verdict could be different. The key point is that TIMERS is using the underlying rule generator in an effective way to produce temporal rule sets that can then be used for determining the verdict.

We reported additional experiments on the weather data (including setting other attributes as the decision attribute) in [12]. In [15] we introduce dependence diagrams, which provide a method for the user to see the amount of influence of each condition attribute on the decision attribute. In the case of soil temperature, the corresponding dependence diagram in [15] shows that the soil temperature depends mainly on its previous value as well as the air temperature.

## 5. Conclusions and Future Work

We have presented a method to discover a set of temporal decision rules that predict or retrodict the value of a decision attribute. Using discovered sets of rules, we can distinguish between instantaneous, possibly causal, and acausal relationships between a decision attribute and a set of condition attributes. Our method relies on the passage of time between the observations of the attribute values.

Possible further work includes the use of aggregate attribute values, which are computed by applying an aggregation function to the values of the condition attributes during a window. For example, the maximum, minimum, or average temperature values during the window are aggregate values that may influence the decision attribute. The definitions of potential causality and acausality could be adjusted to allow for aggregate values.

The sliding position method allows us to refer to combinations of attribute values, some of which appear before the decision attribute while others appear after the decision attribute. Incorporating this possibility into the concepts of reversibility is another possible line of future work.

The TimeSleuth package is available from http://flash.lakeheadu.ca/~kkarimi.

## Acknowledgments

We gratefully acknowledge the Natural Sciences and Engineering Research Council of Canada for providing funding via a Discovery Grant and a Strategic Project Grant awarded to the second author.

# References

[1] C. Antunes, A. Oliveira, Using Context-Free Grammars to Constrain Apriori-based Algorithms for Mining Temporal Association Rules, *Workshop on Temporal Data Mining* (*KDD2002*), Edmonton, Canada, , pp. 11-24, 2002.

[2] D. J. Berndt, J. Clifford, Finding Patterns in Time Series: A Dynamic Programming Approach, *Advances in Knowledge Discovery and Data Mining*. U.M. Fayyad, G. Piatetsky-Shapiro, P. Smyth, et al. (Eds.), AAAI Press/ MIT Press, pp. 229-248, 1996.

[3] L. Breiman, J.H. Friedman, R.A. Olshen, C.J. Stone, *Classification and Regression Trees*, Wadsworth Inc., 1984.

[4] C. Chatfield, *The Analysis of Time Series: An Introduction*, Chapman and Hall, 1989.

[5] D. Freedman, P. Humphreys, *Are There Algorithms that Discover Causal Structure?*, Technical Report 514, Department of Statistics, University of California at Berkeley, 1998.

[6] C. Granger, Investigating Causal Relations by Econometrics Models and Cross-Spectral Methods, *Econometrica 37*, pp. 424-438, 1969.

[7] J.J. Grefenstette, C.L. Ramsey, A.C. Schultz, Learning Sequential Decision Rules Using Simulation Models and Competition, *Machine Learning 5*(*4*), pp. 355-381, 1990.

[8] P. Humphreys, D. Freedman, The Grand Leap, *British Journal of the Philosophy of Science 47*, pp. 113-123, 1996.

[9] K. Karimi, H.J. Hamilton, Logical Decision Rules: Teaching C4.5 to Speak Prolog, *The Second International Conference on Intelligent Data Engineering and Automated Learning* (*IDEAL'2000*), Hong Kong, pp. 85-90, 2000.

[10] K. Karimi, H.J. Hamilton, Discovering Temporal Rules from Temporally Ordered Data, *The Third International Conference on Intelligent Data Engineering and Automated Learning* (*IDEAL'2002*), Manchester, UK, pp. 25-30, 2002.

[11] K. Karimi, H.J. Hamilton, Distinguishing Causal and Acausal Temporal Relations, *The Seventh Pacific-Asia Conference on Knowledge Discovery and Data Mining* (*PAKDD'2003*), Seoul, South Korea, pp. 234-240, 2003.

[12] K. Karimi, H.J. Hamilton, Using TimeSleuth for Discovering Temporal/Causal Rules: A Comparison of Methods, *The Sixteenth Canadian Artificial Intelligence Conference* (*AI'2003*), Halifax, Nova Scotia, Canada, pp. 175-189, 2003.

[13] K. Karimi, H.J. Hamilton, From Temporal Rules to One Dimensional Rules, *Proceedings of the Workshop on Causality and Causal Discovery, Technical Report CS-2004-02*, Kamran Karimi (Ed.). Department of Computer Science, University of Regina, Regina, Saskatchewan, Canada, pp. 30-44, 2004.

[14] K. Karimi, *Discovery of Causality and Acausality from Temporal Sequential Data*, PhD thesis, Department of Computer Science, University of Regina, Regina, Canada, 2005.

[15] K. Karimi, H.J. Hamilton, Using Dependence Diagrams to Summarize Decision Rule Sets, *The Twenty-First Canadian Artificial Intelligence Conference* (*AI'2008*), Windsor, Ontario, Canada, pp. 163-172, 2008.

[16] R.J. Kennett, K.B. Korb, A.E. Nicholson, Seabreeze Prediction Using Bayesian Networks: A Case Study, *Proc. Fifth Pacific-Asia Conference on Knowledge Discovery and Data Mining* (*PAKDD'2001*). Hong Kong, pp. 148-153, 2001.

[17] E.J. Keogh, and M.J. Pazzani, Scaling up Dynamic Time Warping for Data Mining Applications, *The Sixth ACM SIGKDD International Conference on Knowledge Discovery and Data Mining* (*KDD'2000*), pp. 285-289, 2000.

[18] K.B. Korb, C.S. Wallace, In Search of Philosopher's Stone: Remarks on Humphreys and Freedman's Critique of Causal Discovery, *British Journal of the Philosophy of Science 48,* pp. 543-553, 1997.

[19] A.J. Krener, Acausal Realization Theory, Part I; Linear Deterministic Systems. *SIAM Journal on Control and Optimization 25*(*3*), pp. 499-525, 1987.
16